\documentclass[10pt,twocolumn,letterpaper]{article}

\usepackage{cvpr}
\usepackage{times}
\usepackage{epsfig}
\usepackage{graphicx}
\usepackage{amsmath}
\usepackage{amssymb}

\usepackage{times}
\usepackage{epsfig}
\usepackage{graphicx}
\usepackage{amsmath}
\usepackage{amsfonts}
\usepackage{amssymb}
\usepackage{xspace}
\usepackage{caption}
\usepackage{subcaption}
\usepackage{url}

\newcommand{\NJointsEq}{N_J}
\newcommand{\NJoints}{$\NJointsEq{}$}

\newcommand{\NContextEq}{N_c}
\newcommand{\NContext}{$\NContextEq{}$}

\newcommand{\WidthEq}{\mathit{W}}

\newcommand{\HeightEq}{\mathit{H}}

\newcommand{\heatmapEq}{\mathbf{h}}
\newcommand{\heatmap}{$\heatmapEq{}$}

\newcommand{\Weights}{\mathbf{W}}

\newcommand{\xyIt}{$\mathit{(x,y)}$}
\newcommand{\ijIt}{$\mathit{(i,j)}$}
\newcommand{\WHIt}{$\mathit{{W}\times{H}}$}
\newcommand{\argmax}{\textit{argmax}}
\newcommand{\Softmax}{\textit{Softmax}}
\newcommand{\Softargmax}{\textit{Soft-argmax}}

\usepackage[breaklinks=true,bookmarks=false]{hyperref}

\cvprfinalcopy 

\def\cvprPaperID{****} 

\begin{document}

\title{
  Human Pose Regression by Combining Indirect Part Detection and Contextual
  Information
}

\author{Diogo C. Luvizon
\and
Hedi Tabia\\
ETIS Lab., UMR 8051, Universit\'{e} Paris Seine,\\ %
Universit\'{e} Cergy-Pontoise, ENSEA, CNRS.\\
{\tt\small \{diogo.luvizon,  hedi.tabia, picard\}@ensea.fr}\\
\and
David Picard\\
}

\maketitle

\begin{abstract}
  In this paper, we propose an end-to-end trainable regression approach for
  human pose estimation from still images.
  %
  We use the proposed
  \Softargmax{} function to convert feature maps directly to joint
  coordinates, resulting in a fully differentiable framework.
  Our method is able to learn heat maps representations indirectly, without
  additional steps of artificial ground truth generation.
  Consequently, contextual information can be included to the pose predictions
  in a seamless way.
  We evaluated our method on two very challenging datasets, the Leeds Sports
  Poses (LSP) and the MPII Human Pose datasets, reaching the best performance
  among all the existing regression methods and comparable results to the
  state-of-the-art detection based approaches.
\end{abstract}

\section{Introduction}
\label{sec:introduction}

Human %
pose estimation from still images is a hard
task since the human body is strongly articulated, some parts may not be
visible due to occlusions or low quality images, and the visual appearance of
body parts can change significantly from one pose to another.
Classical methods use keypoint detectors to extract local information, which
are combined to build pictorial structures~\cite{Felzenszwalb_IJCV_2005}.
To handle difficult cases of occlusion or partial visualization, contextual
information is usually needed to provide visual cues that can be extracted
from a broad region around the part location~\cite{Fan_2015_CVPR} or by
interaction among detected parts~\cite{Yang_CVPR_2012}.
In general, pose estimation can be seen from two different perspectives,
namely as a correlated part detection problem or as a regression problem.
Detection based approaches commonly try to detect keypoints individually, which
are aggregated in post-processing stages to form one pose prediction.
In contrast, methods based on regression use a function to map directly input
images to body joint positions.


In the last few years, pose estimation have gained attention with the
breakthrough of deep Convolutional Neural Networks
(CNN)~\cite{Toshev_CVPR_2014} alongside consistent computational power
increase.  This can be seen as the shift from classical
approaches~\cite{Pishchulin_ICCV_2013, Ladicky_CVPR_2013} to deep
architectures.
In many recent works from different domains, CNN based methods have overcome
classical approaches by a large margin~\cite{He_2016_CVPR, Simonyan15}.
A key benefit from CNN is that the full pipeline is differentiable, allowing
end-to-end learning.
%
In the context of human pose estimation, the first methods using deep neural
networks tried to do regression directly by learning a non-linear mapping
function from RGB images to joint coordinates~\cite{Toshev_CVPR_2014}.
By contrast, the majority of the methods in the state of the art tackle pose
estimation as a detection problem by predicting heat maps that corresponds to
joint locations~\cite{Newell_ECCV_2016, Bulat_ECCV_2016}.
In such methods, the ground truth is artificially generated from joint
positions, generally as a 2D Gaussian distribution centered on the joint
location, while the context information is implicitly learned by the hidden
convolutional layers.
%

Despite achieving state-of-the-art accuracy on 2D pose estimation,
detection based approaches have some limitations.
For example, such methods relies on additional steps to convert heat
maps to joint positions, usually by applying the \textit{argmax} function,
which is not differentiable, breaking the learning chain on neural networks.
Additionally, the precision of predicted keypoints is proportional to that of
the heat map resolution, which leads the top ranked methods~\cite{Chu_CVPR_17,
Newell_ECCV_2016} to high memory consumption and high computational
requirements.
%

Moreover, regression based methods are conceptually more adapted to 2D and 3D
scenarios and can be used indistinctly on both cases~\cite{Rogez_CVPR_2017,
Sun_2017}.
However, the regression function map is sub-optimally learned, resulting in
lower results if compared to detection based approaches.
In this paper, we aim at solving this problem by bridging the gap between
detection and regression based methods.
We propose to replace the \textit{argmax}
function, used to convert heat maps into joint locations, by what we call the
\Softargmax{} function, which keeps the properties of specialized part
detectors while being fully differentiable.
With this solution, we are able to optimize our model from end-to-end using
regression losses, \ie, from input RGB images to final $(x,y)$ body joint
coordinates.

\begin{figure}[htbp!]
  \centering
  \begin{subfigure}[b]{0.15\textwidth}
    \includegraphics[width=2.7cm]{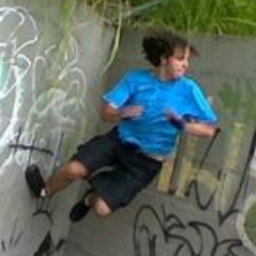}\vspace*{0.2mm}
    \includegraphics[width=2.7cm]{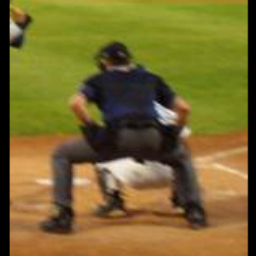}\vspace*{0.2mm}
    \includegraphics[width=2.7cm]{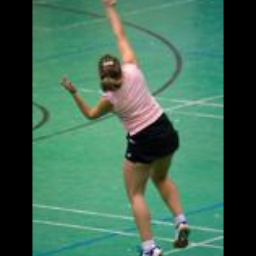}\vspace*{0.2mm}
    \includegraphics[width=2.7cm]{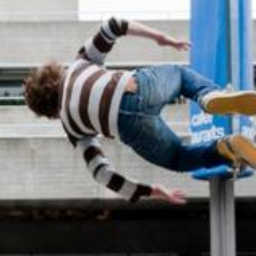}
    \caption{Input image}
  \end{subfigure}
  \hspace{0.2mm}
  \begin{subfigure}[b]{0.15\textwidth}
    \includegraphics[width=2.7cm]{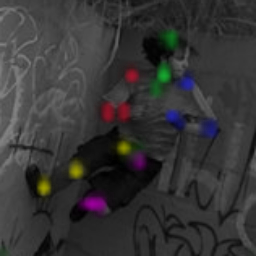}\vspace*{0.2mm}
    \includegraphics[width=2.7cm]{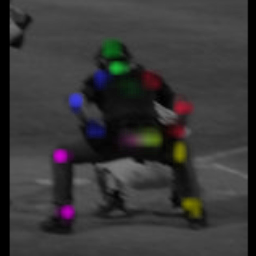}\vspace*{0.2mm}
    \includegraphics[width=2.7cm]{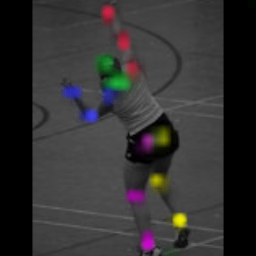}\vspace*{0.2mm}
    \includegraphics[width=2.7cm]{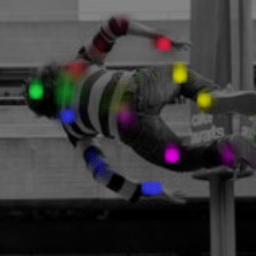}
    \caption{Part-based maps}
  \end{subfigure}
  \hspace{0.2mm}
  \begin{subfigure}[b]{0.15\textwidth}
    \includegraphics[width=2.7cm]{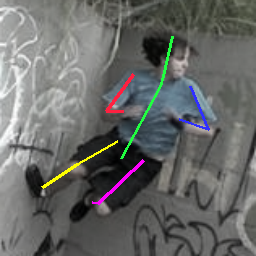}\vspace*{0.2mm}
    \includegraphics[width=2.7cm]{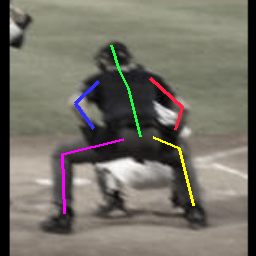}\vspace*{0.2mm}
    \includegraphics[width=2.7cm]{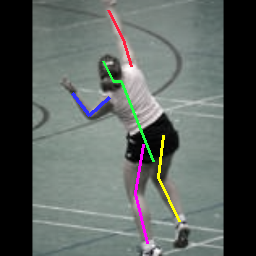}\vspace*{0.2mm}
    \includegraphics[width=2.7cm]{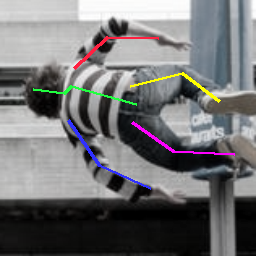}
    \caption{Pose}
  \end{subfigure}
  \caption{
    Test samples from the Leeds Sports Poses (LSP) dataset.
    Input image (a), the predicted part-based maps
    encoded as RGB image for visualizasion (b), and the regressed pose (c).
    Corresponding human limbs have the same colors in all images.
    This figure is better seen in color.
  }
  \label{fig:intro}
\end{figure}

The contributions of our work are the following:
first, we present a human pose regression approach from still images based
on the \Softargmax{} function, resulting in an end-to-end trainable
method which does not require artificial heat maps generation for training.
Second, the proposed method can be trained using an insightful regression loss
function, which is directly linked to the error distance between predicted and
ground truth joint positions.
Third, in the proposed architecture, contextual information is directly
accessible and is easily aggregated to the final predictions.
Finally, the accuracy reached by our method surpasses that of regression
methods ans is close to that of state-of-the-art detection methods, despite
using a much smaller network.
Some examples of our regressed poses are shown in Fig.~\ref{fig:intro}.
Additionally, we provide our implementation of the proposed method in Python
using the open source Keras library~\cite{chollet2015keras}.\footnote{
  The Python source code will be publicly available after acceptance
  at \url{https://github.com/dluvizon/pose-regression}.}

The rest of this paper is divided as follows. In the next section, we present a
review of the most relevant related work. The proposed method is presented in
Section~\ref{sec:method}. In Section~\ref{sec:experiments}, we show the
experimental evaluations, followed by our conclusions in
Section~\ref{sec:conclusion}.

\section{Related work}
\label{sec:related-work}

Several approaches for human pose estimation have been presented for both
2D~\cite{Dantone_CVPR_2013, Lifshitz_ECCV_2016} and 3D~\cite{Ionesc_ICCV_2011}
scenarios, as well as for video sequences~\cite{pfister2015flowing}.
Among classical methods, Pictorial Structures~\cite{Andriluka_CVPR_2009} and
poselet-based features~\cite{Pishchulin_CVPR_2013} have been widely used in
the past.
In this section, due to limited space, we focus on CNN based methods that are
more related to our work.  We briefly refer to the most recent works,
splitting them as regression based and detection based approaches.

\textbf{Regression based approaches.}
%
Some methods tackled pose estimation as a keypoint regression problem.
One of the first regression approaches was proposed by Toshev and
Szegedy~\cite{Toshev_CVPR_2014} as a holistic solution based on cascade
regression for body part detection, where individual joint positions are
recursively improved, taking the full frame as input.
Pfister \etal~\cite{pfister2014deep} proposed the Temporal Pose ConvNet
to track upper body parts, and Carreira \etal~\cite{Carreira_CVPR_2016}
proposed the Iterative Error Feedback by injecting the prediction error back to
the input space, improving estimations recursively.
Recently, Sun \etal~\cite{Sun_2017} proposed a structured bone based
representation for human pose, which is statistically less variant than
absolute joint positions and can be indistinctly used for both 2D and
3D representations.
However, the method requires converting pose data to the relative bone based
format. Moreover, those results are all outperformed by detection based
methods.


\textbf{Detection based approaches.}
Pischulin \etal~\cite{Pishchulin_CVPR_2016} proposed
DeepCut, a graph cutting algorithm that relies on body parts detected by
DeepPose~\cite{Toshev_CVPR_2014}.  This method has been improved
in~\cite{Insafutdinov_ECCV_2016} by replacing the previous CNN by a deep
Residual Network (ResNet)~\cite{He_CVPR_2016}, resulting in very competitive
accuracy results, specially on multi-person detection.
%
More recent methods have shown significant improvements on accuracy by using
fully convolutional models to generate belief maps (or heat maps) for joint
probabilities~\cite{Rafi_BMVC_2016, Belagiannis_ICCV_2015, Bulat_ECCV_2016,
Newell_ECCV_2016, Chu_CVPR_17}.
For example, Bulat \etal~\cite{Bulat_ECCV_2016} proposed a two-stages CNN for
coarse and fine heat map regression using pre-trained models, and
%
%
Gkioxari \etal~\cite{Gkioxari_ECCV_2016} presented a structured prediction
method, where the prediction of each joint depends on the intermediate feature
maps and the distribution probability of the previously predicted joints.
%
%
%
Following the tendency of deeper models with residual connections, Newell
\etal~\cite{Newell_ECCV_2016} proposed a stacked hourglass network with
convolutions in multi-level features, allowing reevaluation of previous
estimations due to a stacked block architecture with many intermediate
supervisions.
The part-based learning process can benefit from intermediate supervision
because it acts as constraints on the lower level layers. As a result,
the feature maps on higher levels tend to be cleaner.
Recently, Chu \etal~\cite{Chu_CVPR_17} extended the stacked hourglass network
by increasing the network complexity and using a CRF based attention map
alongside intermediate supervisions to reinforce the task of learning
structural information.
These results provide to our knowledge state-of-the-art performance.

All the previous methods that are based on detection need additional
steps on training to produce artificial ground truth from joint positions,
which represent an additional processing stage and additional hype-parameters,
since the ground truth heat maps have to be defined by hand.  On evaluation,
the inverse operation is required, \ie, heat maps have to be converted to joint
positions, generally using the $\mathit{argmax}$ function.
Consequently, in order to achieve good precision, predicted heat maps need
reasonable spacial resolution, which increases quadratically the computational
cost and memory usage.
In order to provide an alternative to heat maps based approaches, we
present our framework in the following section.

\section{Proposed method}
\label{sec:method}

The proposed approach is an end-to-end trainable network which
takes as input RGB images and outputs two vectors: the probability
$\mathbf{p}_n$ of joint $n$ being in the image and the regressed joint
coordinates $\mathbf{y}_n = \mathit{(x_n,y_n)}$, where
$n = \{1,2,\dots,\NJointsEq{}\}$ is the index of each joint and \NJoints{} is
the number of joints.
In what follows, we first present the global architecture of our method, and
then detail its most important parts.

\subsection{Network architecture}

\begin{figure*}[htbp]
  \centering
  \includegraphics[width=18.00cm]{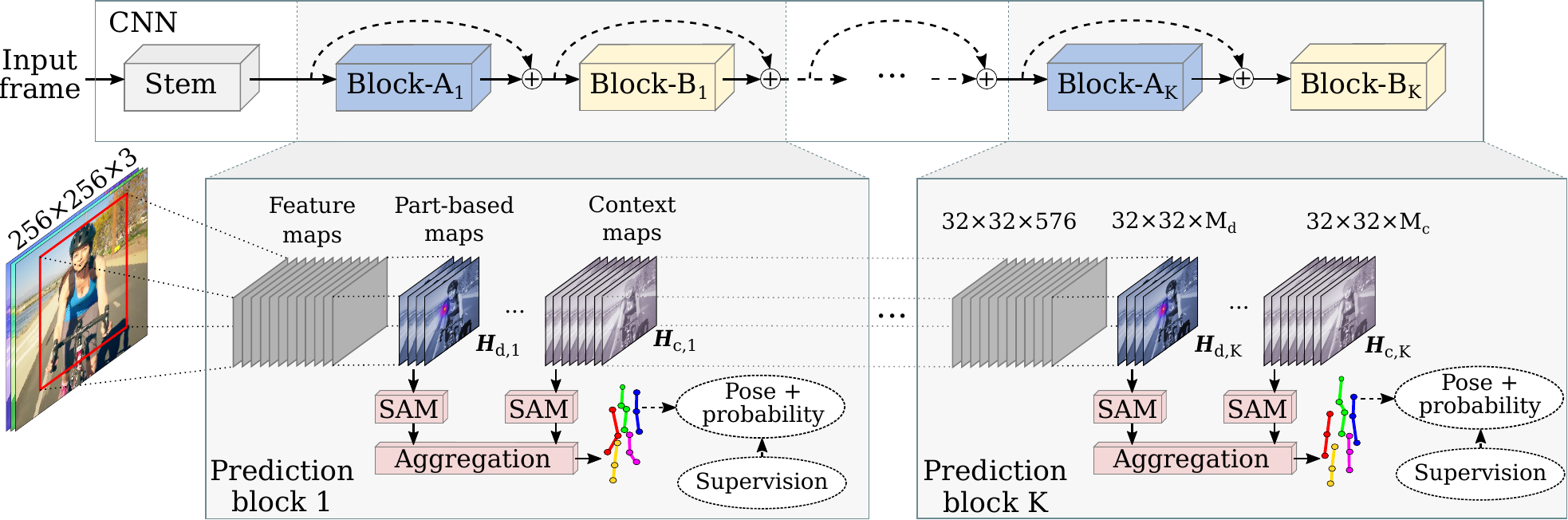}
  \caption{
    Overview of the proposed approach for pose regression.
  }
  \label{fig:netarch}
\end{figure*}

An overview of the proposed method is presented in Fig.~\ref{fig:netarch}.
Our approach is based on a convolutional neural network essentially composed of
three parts: one entry flow (Stem), Block-A and Block-B.
The role of the stem is to provide basic feature extraction, while Block-A
provides refined features and Block-B provides body-part and contextual
activation maps.
One sequence of Block-A and Block-B is used to build one \textit{prediction
block}, which output is used as intermediate supervision during training.
The full network is composed by the stem and a sequence of $K$ prediction
blocks. The final prediction is the output of the $K^{th}$ prediction block.
To predict the pose at each prediction block, we aggregate the 2D coordinates
generated by applying \Softargmax{} to the part-based and contextual maps that
are output by Block-B.
Similarly to recent approaches~\cite{Newell_ECCV_2016, Chu_CVPR_17}, on each
prediction block we produce one estimation that is used as intermediate
supervision, providing better accuracy and more stability to the learning
process.
As a convention, we use the generic term ``heat map'' to refer both to
part-based and contextual feature maps, since these feature maps converge to
heat maps like representations.

The proposed CNN model is partially based on Inception-v4~\cite{SzegedyIV16}
and on the Stacked Hourglass~\cite{Newell_ECCV_2016} networks. %
We also inspired our model on the ``extreme'' inception
(Xception)~\cite{Chollet2016CoRR} network, which relies on the premise that
convolutions can be separated into spatial convolutions (individual for each
channel) followed by a $1\times1$ convolution for cross-channel projection,
resulting in significant reduction on the number of parameters and on
computations.  This idea is called \textit{depthwise separable convolution}
(SepConv) and an optimized implementation is available on the TensorFlow
framework.

\begin{figure}[htbp!]
  \centering
  \begin{subfigure}[b]{0.25\textwidth}
    \centering
    \includegraphics[scale=1.0]{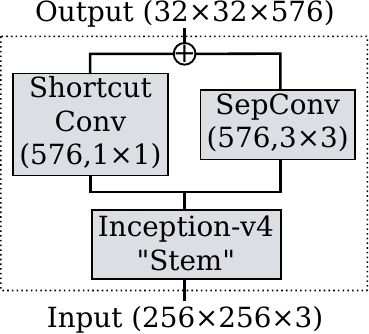}
    \caption{}
    \label{fig:stem-a}
  \end{subfigure}%
  \begin{subfigure}[b]{0.25\textwidth}
    \centering
    \includegraphics[scale=1.0]{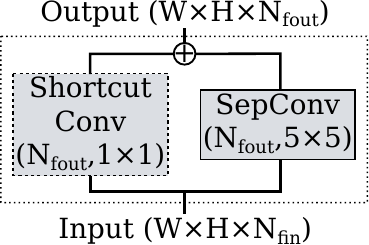}
    \caption{}
    \label{fig:stem-b}
  \end{subfigure}%
  \caption{
    In the proposed network architecture, the Stem (a) is based on
    Inception-v4's stem~\cite{SzegedyIV16} followed by a separable convolution
    in parallel to a shortcut connection. In (b), we present the residual
    separable convolution (Res-SepConv), used to replace the residual block in
    the Stacked Hourglass~\cite{Newell_ECCV_2016} model. If $N_{fin}$ is equal
    to $N_{fout}$, the shortcut convolution is replaced by the identity
    mapping.
  }
  \label{fig:stem}
\end{figure}

In our network, the ``Stem'' is based on Inception-v4's stem followed by a SepConv
layer in parallel with a shortcut layer, as presented in
Fig.~\ref{fig:stem-a}. %
For Block-A, we use a similar architecture as the Stacked Hourglass replacing
all the residual blocks by a residual separable convolution (Res-SepConv), as
depicted in Fig.~\ref{fig:stem-b}. %
Additionally, our approach increased the results from~\cite{Newell_ECCV_2016}
with only three feature map resolutions, from $32\times32$ to $8\times8$,
instead of the original five resolutions, from $64\times64$ to $4\times4$.

\begin{figure}[htbp]
  \centering
  \begin{subfigure}[b]{0.5\textwidth}
    \centering
    \includegraphics[scale=1.0]{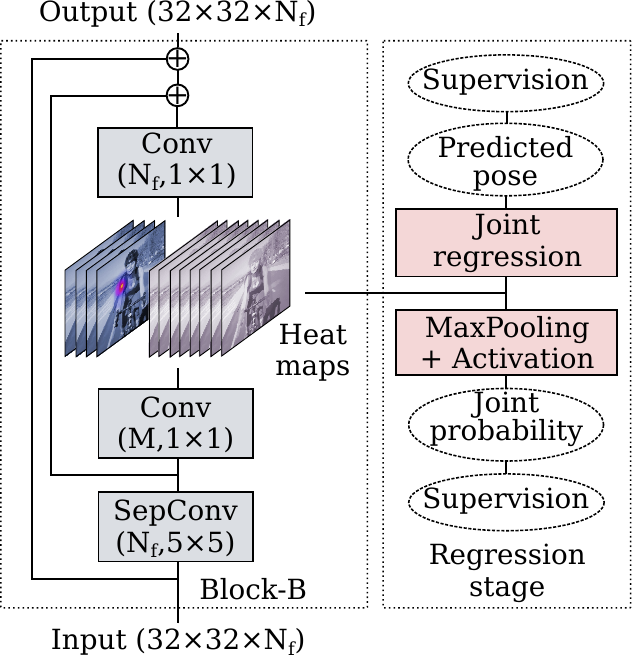}
  \end{subfigure}%
  \caption{
    Network architecture of Block-B and an overview of the regression stage.
    The input is projected into $M$ heat maps ($M_d + M_c$ ) which are then
    used for pose regression.
  }
  \label{fig:blockb}
\end{figure}

At each prediction stage, Block-B is used to transform input feature maps into
$M_d$ \textit{part-based detection maps} ($\mathbf{H}_d$) and $M_c$
\textit{context maps} ($\mathbf{H}_c$), resulting in $M = M_d + M_c$ heat maps.
For the specific problem of pose estimation, $M_d$ corresponds to the number
of joints \NJoints{}, and $M_c = \NContextEq{}\NJointsEq{}$, where \NContext{}
is the number of context maps for each joint.
%
%
The produced heat maps are projected back to the feature space and reintroduced
to the network flow by a $1\times1$ convolution.  Similar techniques have been
used by many previous works~\cite{Bulat_ECCV_2016, Newell_ECCV_2016,
Chu_CVPR_17}, resulting in significant gain of performance.
From the generated heat maps, our method computes the predicted joint locations
and joint probabilities in the regression block, which has no trainable
parameters.
The architecture of Block-B and the regression stage is shown in
Fig.~\ref{fig:blockb}.

\subsection{Proposed regression method}

As presented in Section~\ref{sec:related-work}, traditional regression based
methods use fully connected layers on feature maps and learn the regression
mapping.  However, this approach usually gives sub-optimal solutions.
While state-of-the-art methods are overwhelmingly based on part detection,
approaches based on regression have the advantages of providing directly the
pose prediction as joint coordinates without additional steps or
post-processing.
In order to provide an alternative to detection based methods, we propose an
efficient and fully differentiable way to convert heat maps directly to \xyIt{}
coordinates, which we call \Softargmax{}.
Additionally, the \Softargmax{} operation can be implemented as a CNN layer,
as detailed in the next section.

\subsubsection{Soft-argmax layer}
\label{sec:sam}

Let us define the \Softmax{} operation on a single heat map $\heatmapEq{} \in
\mathbb{R}^{W\times{H}}$ as:
\begin{equation}
  \Phi(\heatmapEq{}_{i,j}) = \frac{e^{\heatmapEq{}_{i,j}}}{%
    \sum_{k=1}^{\WidthEq{}}%
    \sum_{l=1}^{\HeightEq{}}%
    e^{\heatmapEq{}_{k,l}}
    },
  \label{eq:softmax}
\end{equation}
where $\heatmapEq_{i,j}$ is the value of heat map $\heatmapEq$ at location
\ijIt{}, and \WHIt{} is the heat map size.
Contrary to the more common cross-channel softmax, we use here a spatial
softmax that ensures each heat maps is normalized.
Then, we define the \Softargmax{} as follows:
\begin{equation}
  \Psi_{d}(\heatmapEq{}) = \sum_{i=1}^{\WidthEq{}}\sum_{j=1}^{\HeightEq{}}%
  \Weights{}_{i,j,d}\Phi(\heatmapEq{}_{i,j}),
  \label{eq:softargmax-math}
\end{equation}
where $\mathit{d}$ is a given component $\mathit{x}$ or $\mathit{y}$, and
$\Weights{}$ is a $\WidthEq{}\times\HeightEq{}\times{2}$ weight matrix
corresponding to the coordinates \xyIt{}.
The matrix $\Weights{}$ can be expressed by its components
$\Weights{}_{x}$ and $\Weights{}_{y}$, which are 2D discrete normalized ramps,
defined as follows:
\begin{equation}
  \Weights{}_{i,j,x} =  \frac{i}{W}, \Weights{}_{i,j,y} = \frac{j}{H}.
  \label{eq:weights-argmax}
\end{equation}
Finally, given a heat map \heatmap{}, the regressed location of the predicted
joint is given by
\begin{equation}
  \mathbf{y} = (\Psi_{x}(\heatmapEq{}), \Psi_{y}(\heatmapEq{}))^T.
  \label{eq:regression}
\end{equation}

This \Softargmax{} operation can be seen as a weighted average of points
distributed on an uniform grid, with the weights being equal to the
corresponding heat map.

In order to integrate the \Softargmax{} layer into a deep network, we need its
derivative with respect to \heatmap{}:
\begin{equation}
  \frac{\partial \Psi_{d}(\heatmapEq{}_{i,j})}{\partial \heatmapEq{}_{i,j}} = %
    \Weights{}_{i,j,d} %
    \frac{e^{\heatmapEq{}_{i,j}} %
      (\sum_{k=1}^{\WidthEq{}} \sum_{l=1}^{\HeightEq{}}e^{\heatmapEq{}_{k,l}} - e^{\heatmapEq{}_{i,j}}) %
    }{
      (\sum_{k=1}^{\WidthEq{}} \sum_{l=1}^{\HeightEq{}}e^{\heatmapEq{}_{k,l}})^2}.
  \label{eq:derivative}
\end{equation}
The \Softargmax{} function can thus be integrated in a trainable framework by
using back propagation and the chain rule on equation (\ref{eq:derivative}).
Moreover, from equation (\ref{eq:derivative}), we can see that the gradient is
exponentially increasing for higher values, resulting in very discriminative
response at the joint position.

The implementation of \Softargmax{} can be easily done with recent frameworks,
such as TensorFlow, just by concatenating a spatial softmax followed by
one convolutional layer with 2 filters of size $\WidthEq{}\times\HeightEq{}$,
with fixed parameters according to equation~(\ref{eq:weights-argmax}).

Unlike traditional \argmax{}, \Softargmax{} provides sub-pixel accuracy,
allowing good precision even with very low resolution.
Moreover, the \Softargmax{} operation allows to learn very discriminative
heat maps directly from the \xyIt{} joint coordinates without explicitly
computing artificial ground truth.
Samples of heat maps learned by our approach are shown in
Fig.~\ref{fig:part-based-maps}.

\subsubsection{Joint probability}

Additionally to the joint locations, we estimate the joint probability
$\mathbf{p}_n$, which corresponds to the probability of the $n^{th}$ joint
being present in the image.
The estimated joint probability is given by the sigmoid activation on the
global max-pooling from heat map $\heatmapEq{}_{n}$.
Despite giving an additional piece of information, the joint probability does
not depends on additional parameters and is computationally negligible,
compared to the cost of convolutional layers.

\subsubsection{Detection and context aggregation}

Even if the correlation between some joints can be learned in the hidden
convolutional layers, the joint regression approach is designed to
locate body parts individually, resulting in low flexibility to learn from the
context.
For example, the same filters that give high response to images of a clean
head, also must react positively to a hat or a pair of sunglasses.
In order to provide multi-source information to the final prediction, we
include in our framework specialized part-based heat maps and context heat
maps, which are defined as
$\mathbf{H}_d = [\mathbf{h}^d_1, \dots, \mathbf{h}^d_{\NJointsEq{}}]$ and
$\mathbf{H}_c = [\mathbf{h}^c_{1,1}, \dots, \mathbf{h}^c_{N_c, N_j}]$,
respectively.
Additionally, we define the joint probability related to each context map
as $\mathbf{p}^{c}_{i,n}$, where $i = \{1, \dots, N_c\}$ and
$n = \{1, \dots, N_j\}$.

Finally, the $n^{th}$ joint position from detection and contextual information
aggregated is given by:
\begin{equation}
  \mathbf{y}_n = \alpha{\mathbf{y}^d_n} +
  (1 - \alpha)\frac{\sum_{i=1}^{N_c}\mathbf{p}^c_{i,n}\mathbf{y}^c_{i,n}}{\sum_{i=1}^{N_c}\mathbf{p}^c_{i,n}},
  \label{eq:aggregation}
\end{equation}
where $\mathbf{y}^d_n = \Softargmax{}(\mathbf{h}^d_n)$ is the predicted
location from the $n^{th}$ part based heat map,
$\mathbf{y}^c_{i,n} = \Softargmax{}(\mathbf{h}^c_{i,n})$ and
$\mathbf{p}^c_{i,n}$ are respectively the location and the probability for the
$i^{th}$ context heat map for joint $n$, and $\alpha$ is a hyper-parameter.

From equation (\ref{eq:aggregation}) we can see that the final prediction
is a combination of one specialized prediction and $N_c$ contextual
predictions pondered by their probabilities.
The contextual weighted contribution brings flexibility, allowing specific
filters to be more responsive to particular patterns.
This aggregation scheme within the learning stage is only possible because we
have the joint probability and position directly available inside the network
in a differentiable way.

\section{Experiments}
\label{sec:experiments}

We evaluate the proposed method on the very challenging MPII Human
Pose~\cite{Andriluka_CVPR_2014} and Leeds Sports Poses
(LSP)~\cite{Johnson_BMVC_2010} datasets.
The MPII dataset contains 25K images collected from YouTube videos, including
around 28K annotated poses for training and 15K poses for testing.
The annotated poses have 16 body joints, some of them are not present and
others are occluded but can be predicted by the context.
The LSP dataset is composed by 2000 annotated poses with up to 14 joint
locations.  The images were gethered from Flickr with sports people.
The details about training the model and achieved accuracy results are given as
follows.

\subsection{Training}


The proposed network was trained simultaneously on joints regression
and joint probabilities.  For joints regression, we use the elastic net loss
function (L1 + L2):
\begin{equation}
  \mathit{L}_{\mathbf{y}} = \frac{1}{\NJointsEq{}}%
  \sum_{n=1}^{\NJointsEq{}}%
  \|\mathbf{y}_n - \hat{\mathbf{y}}_n\|_1 +%
  \|\mathbf{y}_n - \hat{\mathbf{y}}_n\|_2^2,
  \label{eq:l1l2-loss}
\end{equation}
where $\mathbf{y}_n$ and $\hat{\mathbf{y}}_n$ are respectively the ground truth
and the predicted $n^{th}$ joint coordinates.
In this case, we use directly the joint coordinates normalized to the interval
$[0,1]$, where the top-left image corner corresponds to $(0,0)$, and the
bottom-right image corner corresponds to $(1,1)$.

For joint probability estimation, we use the binary cross entropy loss
function on the joint probability $\mathbf{p}$:
\begin{equation}
  \mathit{L}_{\mathbf{p}} = \frac{1}{\NJointsEq{}}%
  \sum_{n=1}^{\NJointsEq{}}%
  [(\mathbf{p}_n-1)~log~(1-\hat{\mathbf{p}}_n)%
  -\mathbf{p}_n~log~\hat{\mathbf{p}}_n],
  \label{eq:loss-prob}
\end{equation}
where $\mathbf{p}_n$ and $\hat{\mathbf{p}}_n$ are respectively the ground
truth and the predicted joint probability.

We optimize the network using back propagation and the RMSProp optimizer,
with batch size of 16 samples.
%
%
For the MPII dataset, we train the network for 120 epochs.  The learning rate
begins at $10^{-3}$ and decreases by a factor of $0.4$ when accuracy on
validation plateaus.
We use the same validation split as proposed in~\cite{Tompson_CVPR_2015}.
On the LSP dataset, we start from the model trained on MPII and fine-tuned it
for more 70 epochs, beginning with learning rate $2\cdot10^{-5}$ and using the
same decrease procedure.
The full training of our network takes three days on the relatively outdated
NVIDIA GPU Tesla K20 with 5GB of memory.

\textbf{Data augmentation.}
We used standard data augmentation on both MPII and LSP datasets. Input RGB
images were cropped and centered on the main subject with a squared bounding
box, keeping the people scale (when provided), then resized to $256\times256$
pixels.  We perform random rotations ($\pm40^{\circ}$) and random rescaling
from $0.7$ to $1.3$ on MPII and from $0.85$ to $1.25$ on LSP to make the model
more robust to image changes.

\textbf{Parameters setup.}
The network model is defined according to Fig.~\ref{fig:netarch} and composed
of eight prediction blocks ($K=8$).
We trained the network to regress 16 joints with 2 context maps for each joint
($N_j = 16$, $\NContextEq{} = 2$).
In the aggregation stage, we use $\alpha=0.8$.

\subsection{Results}

\textbf{LSP dataset.}
We evaluate our method on the LSP dataset using two metrics, the ``Percentage
of Correct Parts'' (PCP) and the ``Probability of Correct Keypoint'' (PCK)
measures, as well as two different evaluation protocols, ``Observer-Centric``
(OC) and ``Person-Centric`` (PC), resulting in four different evaluation
settings.
Our results compared to the state-of-the-art results on the LSP dataset using
OC annotations are present in Table~\ref{tab:lsp-pck-oc} (PCK measure) and
Table~\ref{tab:lsp-pcp-oc} (PCP measure).
In both cases, we overcome the best scores by a significant margin, specially
with respect to the lower leg and the ankles, on which we increase the results
of Pishchulin \etal~\cite{Pishchulin_CVPR_2016}
by 6.3\% and 4.6\%, respectively.
To the best of our knowledge, we are the sole regression method to report
results using this evaluation settings.

\begin{table}[]
\centering
\caption{Results on LSP test samples using the PCK measure at 0.2 with OC annotations.}
\label{tab:lsp-pck-oc}
\resizebox{0.5\textwidth}{!}{%
\begin{tabular}{@{}lcccccccc@{}}
    \hline
    Method & Head & Sho. & Elb. & Wri. & Hip & Knee & Ank. & PCK \\ \hline
    \multicolumn{9}{c}{Detection based methods} \\
    Kiefel and Gehler~\cite{Kiefel2014} & 83.5  & 73.7  & 55.9  & 36.2  & 73.7  & 70.5 & 66.9 & 65.8  \\
    Ramakrishna \etal~\cite{Ramakrishna_ECCV_2014} & 84.9  & 77.8  & 61.4  & 47.2  & 73.6  & 69.1 & 68.8 & 69.0 \\
    Pishchulin \etal~\cite{Pishchulin_ICCV_2013} & 87.5  & 77.6  & 61.4  & 47.6  & 79.0  & 75.2 & 68.4 & 71.0 \\
    Ouyang \etal~\cite{Ouyang_CVPR_2014} & 86.5  & 78.2  & 61.7  & 49.3  & 76.9  & 70.0 & 67.6 & 70.0 \\
    Chen and Yuille~\cite{Chen_NIPS14} & 91.5  & 84.7  & 70.3  & 63.2  & 82.7  & 78.1 & 72.0 & 77.5 \\
    Yang \etal~\cite{yang2016end} & 90.6  & 89.1  & 80.3  & 73.5  & 85.5  & 82.8 & 68.8 & 81.5 \\
    Chu \etal~\cite{chu2016structure} & 93.7  & 87.2  & 78.2  & 73.8  & 88.2  & 83.0 & 80.9 & 83.6 \\
    Pishchulin \etal~\cite{Pishchulin_CVPR_2016} & \textbf{97.4}  & 92.0  & 83.8  & 79.0  & 93.1  & 88.3 & 83.7 & 88.2 \\ \hline
    \multicolumn{9}{c}{Regression based method} \\
    Our method & \textbf{97.4}  & \textbf{93.8}  & \textbf{86.8}  & \textbf{82.3}  & \textbf{93.7}  & \textbf{90.9} & \textbf{88.3} & \textbf{90.5} \\ \hline
\end{tabular}}
\end{table}

\begin{table}[]
\centering
\caption{Results on LSP test samples using the PCP measure with OC annotations.}
\label{tab:lsp-pcp-oc}
\resizebox{0.5\textwidth}{!}{%
\begin{tabular}{@{}lccccccc@{}}
    \hline
    Method & Torso & Upper & Lower & Upper & Fore- & Head & PCP \\
           &       & leg   & leg   & arm   & arm   &      &     \\ \hline
    \multicolumn{8}{c}{Detection based methods} \\
    Kiefel and Gehler~\cite{Kiefel2014} & 84.3  & 74.5  & 67.6  & 54.1  & 28.3  & 78.3 & 61.2 \\
    Pishchulin \etal~\cite{Pishchulin_CVPR_2013} &  87.4  & 75.7  & 68.0  & 54.4  & 33.7  & 77.4 & 62.8 \\
    Ramakrishna \etal~\cite{Ramakrishna_ECCV_2014} & 88.1  & 79.0  & 73.6  & 62.8  & 39.5  & 80.4 & 67.8 \\
    Ouyang \etal~\cite{Ouyang_CVPR_2014} & 88.6  & 77.8  & 71.9  & 61.9  & 45.4  & 84.3 & 68.7 \\
    Pishchulin \etal~\cite{Pishchulin_ICCV_2013} & 88.7  & 78.9  & 73.2  & 61.8  & 45.0  & 85.1 & 69.2 \\
    Chen and Yuille~\cite{Chen_NIPS14} & 92.7  & 82.9  & 77.0  & 69.2  & 55.4  & 87.8 & 75.0 \\
    Yang \etal~\cite{yang2016end} & 96.5  & 88.7  & 81.7  & 78.8  & 66.7  & 83.1 & 81.1 \\
    Chu \etal~\cite{chu2016structure} & 95.4  & 87.6  & 83.2  & 76.9  & 65.2  & 89.6 & 81.1 \\
    Pishchulin \etal~\cite{Pishchulin_CVPR_2016} & 96.0  & 91.0  & 83.5  & 82.8  & 71.8  & \textbf{96.2} & 85.0 \\ \hline
    \multicolumn{8}{c}{Regression based method} \\
    Our method & \textbf{98.2}  & \textbf{93.8}  & \textbf{89.8}  & \textbf{85.8}  & \textbf{75.5}  & 96.0 & \textbf{88.4} \\ \hline
\end{tabular}}
\end{table}

Using the PC annotations on LSP, we achieved the best results among regression
based approaches and the second general score, as presented in
Tables~\ref{tab:lsp-pck-pc} and~\ref{tab:lsp-pcp-pc}.
On the PCK measure, we outperform the results reported by Carreira
\etal~\cite{Carreira_CVPR_2016} (CVPR 2016), which is the only regression
method reported on this setup, by 18.0\%.

\begin{table}[]
\centering
\caption{Results on LSP test samples using the PCK measure at 0.2 with PC annotations.}
\label{tab:lsp-pck-pc}
\resizebox{0.5\textwidth}{!}{%
\begin{tabular}{@{}lcccccccc@{}}
    \hline
    Method & Head & Sho. & Elb. & Wri. & Hip & Knee & Ank. & PCK \\ \hline
    \multicolumn{9}{c}{Detection based methods} \\
    Pishchulin \etal \cite{Pishchulin_ICCV_2013} & 87.2  & 56.7  & 46.7  & 38.0  & 61.0  & 57.5 & 52.7 & 57.1 \\
    Chen and Yuille~\cite{Chen_NIPS14} & 91.8  & 78.2  & 71.8  & 65.5  & 73.3  & 70.2 & 63.4 & 73.4 \\
    Fan \etal \cite{Fan_2015_CVPR} & 92.4  & 75.2  & 65.3  & 64.0  & 75.7  & 68.3 & 70.4 & 73.0 \\
    Tompson \etal \cite{Tompson_NIPS_2014} & 90.6  & 79.2  & 67.9  & 63.4  & 69.5  & 71.0 & 64.2 & 72.3 \\
    Yang \etal \cite{yang2016end}  & 90.6  & 78.1  & 73.8  & 68.8  & 74.8  & 69.9 & 58.9 & 73.6 \\
    Rafi \etal \cite{Rafi_BMVC_2016} & 95.8  & 86.2  & 79.3  & 75.0  & 86.6  & 83.8 & 79.8 & 83.8 \\
    Yu \etal \cite{Yu2016} & 87.2  & 88.2  & 82.4  & 76.3  & 91.4  & 85.8 & 78.7 & 84.3 \\
    Belag. and Ziss. \cite{Belagiannis_2016} & 95.2  & 89.0  & 81.5  & 77.0  & 83.7  & 87.0 & 82.8 & 85.2 \\
    Lifshitz \etal \cite{Lifshitz_ECCV_2016} & 96.8  & 89.0  & 82.7  & 79.1  & 90.9  & 86.0 & 82.5 & 86.7 \\
    Pishchulin \etal \cite{Pishchulin_CVPR_2016} & 97.0  & 91.0  & 83.8  & 78.1  & 91.0  & 86.7 & 82.0 & 87.1 \\
    Insafutdinov \etal \cite{Insafutdinov_ECCV_2016} & 97.4  & 92.7  & 87.5  & 84.4  & 91.5  & 89.9 & 87.2 & 90.1 \\
    Wei \etal \cite{Wei_CVPR_2016} & 97.8  & 92.5  & 87.0  & 83.9  & 91.5  & 90.8 & 89.9 & 90.5 \\
    Bulat and Tzimi. \cite{Bulat_ECCV_2016} & 97.2  & 92.1  & 88.1  & 85.2  & 92.2  & 91.4 & 88.7 & 90.7 \\
    Chu \etal \cite{Chu_CVPR_17} & \textbf{98.1}  & \textbf{93.7}  & \textbf{89.3}  & \textbf{86.9}  & \textbf{93.4}  & \textbf{94.0} & \textbf{92.5} & \textbf{92.6} \\ \hline
    \multicolumn{9}{c}{Regression based methods} \\
    Carreira \etal \cite{Carreira_CVPR_2016} & 90.5  & 81.8  & 65.8  & 59.8  & 81.6  & 70.6 & 62.0 & 73.1 \\
    Our method & \textbf{97.5}  & \textbf{93.3}  & \textbf{87.6}  & \textbf{84.6}  & \textbf{92.8}  & \textbf{92.0} & \textbf{90.0} & \textbf{91.1} \\ \hline
\end{tabular}}
\end{table}

\begin{table}[]
\centering
\caption{Results on LSP test samples using the PCP measure with PC annotations.}
\label{tab:lsp-pcp-pc}
\resizebox{0.5\textwidth}{!}{%
\begin{tabular}{@{}lccccccc@{}}
    \hline
    Method & Torso & Upper & Lower & Upper & Fore- & Head & PCP \\
           &       & leg   & leg   & arm   & arm   &      &     \\ \hline
    \multicolumn{8}{c}{Detection based methods} \\
    Pishchulin \etal \cite{Pishchulin_ICCV_2013} & 88.7  & 63.6  & 58.4  & 46.0  & 35.2  & 85.1 & 58.0 \\
    Tompson \etal \cite{Tompson_NIPS_2014} & 90.3  & 70.4  & 61.1  & 63.0  & 51.2  & 83.7 & 66.6 \\
    Fan \etal \cite{Fan_2015_CVPR} & 95.4  & 77.7  & 69.8  & 62.8  & 49.1  & 86.6 & 70.1 \\
    Chen and Yuille~\cite{Chen_NIPS14} & 96.0  & 77.2  & 72.2  & 69.7  & 58.1  & 85.6 & 73.6 \\
    Yang \etal \cite{yang2016end} & 95.6  & 78.5  & 71.8  & 72.2  & 61.8  & 83.9 & 74.8 \\
    Rafi \etal \cite{Rafi_BMVC_2016} & 97.6  & 87.3  & 80.2  & 76.8  & 66.2  & 93.3 & 81.2 \\
    Belag. and Ziss. \cite{Belagiannis_2016} & 96.0  & 86.7  & 82.2  & 79.4  & 69.4  & 89.4 & 82.1 \\
    Yu \etal \cite{Yu2016} & 98.0  & 93.1  & 88.1  & 82.9  & 72.6  & 83.0 & 85.4 \\
    Lifshitz \etal \cite{Lifshitz_ECCV_2016} & 97.3  & 88.8  & 84.4  & 80.6  & 71.4  & 94.8 & 84.3 \\
    Pishchulin \etal \cite{Pishchulin_CVPR_2016} & 97.0  & 88.8  & 82.0  & 82.4  & 71.8  & 95.8 & 84.3 \\
    Insafutdinov \etal \cite{Insafutdinov_ECCV_2016} & 97.0  & 90.6  & 86.9  & 86.1  & 79.5  & 95.4 & 87.8 \\
    Wei \etal \cite{Wei_CVPR_2016} & 98.0  & 92.2  & 89.1  & 85.8  & 77.9  & 95.0 & 88.3 \\
    Bulat and Tzimi. \cite{Bulat_ECCV_2016} & 97.7  & 92.4  & 89.3  & 86.7  & 79.7  & 95.2 & 88.9 \\
    Chu \etal \cite{Chu_CVPR_17} & \textbf{98.4}  & \textbf{95.0}  & \textbf{92.8}  & \textbf{88.5}  & \textbf{81.2}  & 95.7 & \textbf{90.9} \\ \hline
    \multicolumn{8}{c}{Regression based methods} \\
    Carreira \etal \cite{Carreira_CVPR_2016} & 95.3  & 81.8  & 73.3  & 66.7  & 51.0  & 84.4 & 72.5 \\
    Our method & \textbf{98.2}  & \textbf{93.6}  & \textbf{91.0}  & \textbf{86.6}  & \textbf{78.2}  & \textbf{96.8} & \textbf{89.4} \\ \hline
\end{tabular}}
\end{table}

\begin{table*}[]
  \centering
  \caption{Comparison results with state-of-the-art methods on the MPII dataset
  on testing, using PCKh measure with threshold as 0.5 of the head segment length.
  Detection based methods are shown on top and regression based methods on
  bottom.}
  \label{tab:mpii}
  \begin{tabular}{@{}lcccccccc@{}}
  \hline
  Method & Head & Shoulder & Elbow & Wrist & Hip & Knee & Ankle & Total \\ \hline
  \multicolumn{9}{c}{Detection based methods}  \\
  Pishchulin \etal \cite{Pishchulin_ICCV_2013}      & 74.3  & 49.0  & 40.8  & 34.1  & 36.5  & 34.4 & 35.2 & 44.1 \\
  Tompson \etal \cite{Tompson_NIPS_2014}            & 95.8  & 90.3  & 80.5  & 74.3  & 77.6  & 69.7 & 62.8 & 79.6 \\
  Tompson \etal \cite{Tompson_CVPR_2015}            & 96.1  & 91.9  & 83.9  & 77.8  & 80.9  & 72.3 & 64.8 & 82.0 \\
  Hu and Ramanan \cite{Hu_CVPR_2016}                & 95.0  & 91.6  & 83.0  & 76.6  & 81.9  & 74.5 & 69.5 & 82.4 \\
  Pishchulin \etal \cite{Pishchulin_CVPR_2016}      & 94.1  & 90.2  & 83.4  & 77.3  & 82.6  & 75.7 & 68.6 & 82.4 \\
  Lifshitz \etal \cite{Lifshitz_ECCV_2016}          & 97.8  & 93.3  & 85.7  & 80.4  & 85.3  & 76.6 & 70.2 & 85.0 \\
  Gkioxary \etal \cite{Gkioxari_ECCV_2016}          & 96.2  & 93.1  & 86.7  & 82.1  & 85.2  & 81.4 & 74.1 & 86.1 \\
  Rafi \etal \cite{Rafi_BMVC_2016}                  & 97.2  & 93.9  & 86.4  & 81.3  & 86.8  & 80.6 & 73.4 & 86.3 \\
  Belagiannis and Zisserman \cite{Belagiannis_2016} & 97.7  & 95.0  & 88.2  & 83.0  & 87.9  & 82.6 & 78.4 & 88.1 \\
  Insafutdinov \etal \cite{Insafutdinov_ECCV_2016}  & 96.8  & 95.2  & 89.3  & 84.4  & 88.4  & 83.4 & 78.0 & 88.5 \\
  Wei \etal \cite{Wei_CVPR_2016}                    & 97.8  & 95.0  & 88.7  & 84.0  & 88.4  & 82.8 & 79.4 & 88.5 \\
  Bulat and Tzimiropoulos \cite{Bulat_ECCV_2016}    & 97.9  & 95.1  & 89.9  & 85.3  & 89.4  & 85.7 & 81.7 & 89.7 \\
  Newell \etal \cite{Newell_ECCV_2016}              & 98.2  & 96.3  & 91.2  & 87.1  & 90.1  & 87.4 & 83.6 & 90.9 \\
  Chu \etal \cite{Chu_CVPR_17}                      & \textbf{98.5}  & 96.3  & 91.9  & 88.1  & 90.6  & 88.0 & 85.0 & 91.5 \\
    Chou \etal \cite{ChouCC17}                        & 98.2  & \textbf{96.8}  & 92.2  & 88.0  & \textbf{91.3}  & 89.1 & 84.9 & 91.8 \\
    Chen \etal \cite{ChenSWLY17}                      & 98.1  & 96.5  & \textbf{92.5}  & \textbf{88.5}  & 90.2  & \textbf{89.6} & \textbf{86.0} & \textbf{91.9} \\
    \hline
  \multicolumn{9}{c}{Regression based methods} \\
    Rogez \etal \cite{Rogez_CVPR_2017}              & --    & --    & --    & --    & --    & --   & --   & 74.2 \\
    Carreira \etal \cite{Carreira_CVPR_2016}        & 95.7  & 91.7  & 81.7  & 72.4  & 82.8  & 73.2 & 66.4 & 81.3 \\
    Sun \etal \cite{Sun_2017}                       & 97.5  & 94.3  & 87.0  & 81.2  & 86.5  & 78.5 & 75.4 & 86.4 \\
    Our method                                      & \textbf{98.1}  & \textbf{96.6}  & \textbf{92.0}  & \textbf{87.5}  & \textbf{90.6}  & \textbf{88.0} & \textbf{82.7} & \textbf{91.2} \\ \hline
  \end{tabular}
\end{table*}

\textbf{MPII dataset.}
On the MPII dataset, we evaluate our method using the ``Single person''
challenge~\cite{Andriluka_CVPR_2014}.
The scores were computed by the providers of the dataset, since the test labels
are not publicly available.
As shown in Table~\ref{tab:mpii}, we reached a test score of 91.2\%, which is
only 0.7\% lower then the best result using detection based method, and 4.8\%
higher than the second score using regression.

Taking into account the competitiveness of the MPII Human Pose
challenge\footnote{MPII Leader Board: \url{http://human-pose.mpi-inf.mpg.de}},
our score represents a very significant improvement over regression based
approaches and a promising result compared to detection based methods.
Moreover, our method is much simpler than the stacked hourglass network from
Newell \etal~\cite{Newell_ECCV_2016} or its extension from
Chu \etal~\cite{Chu_CVPR_17}.  Due to limited memory resources, we were not
able to train these two models in our hardware.
Despite that, we reach comparable results with a model that fits in much
smaller GPUs.

\subsection{Discussion}

\begin{figure}[htbp!]
  \centering
  \begin{subfigure}[b]{0.23\textwidth}
    \centering
    \includegraphics[width=3.99cm]{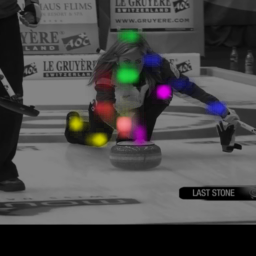}\vspace{0.1mm}\\
    \includegraphics[width=1.96cm]{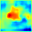}
    \includegraphics[width=1.96cm]{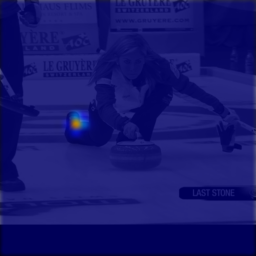}\vspace{0.1mm}\\
    \includegraphics[width=1.96cm]{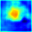}
    \includegraphics[width=1.96cm]{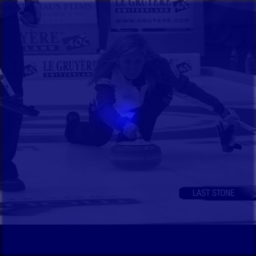}\vspace{0.1mm}\\
    \includegraphics[width=1.96cm]{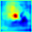}
    \includegraphics[width=1.96cm]{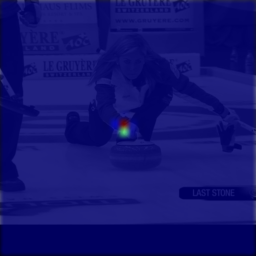}\vspace{0.1mm}\\
    \includegraphics[width=1.96cm]{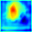}
    \includegraphics[width=1.96cm]{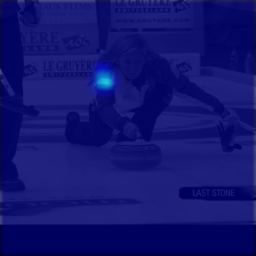}
  \end{subfigure}
  \hspace{0.1mm}
  \begin{subfigure}[b]{0.23\textwidth}
    \centering
    \includegraphics[width=1.96cm]{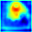}
    \includegraphics[width=1.96cm]{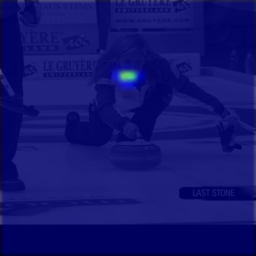}\vspace{0.1mm}\\
    \includegraphics[width=1.96cm]{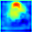}
    \includegraphics[width=1.96cm]{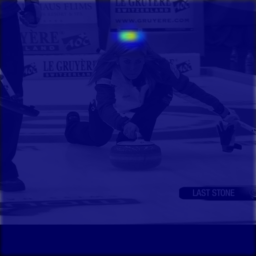}\vspace{0.1mm}\\
    \includegraphics[width=1.96cm]{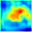}
    \includegraphics[width=1.96cm]{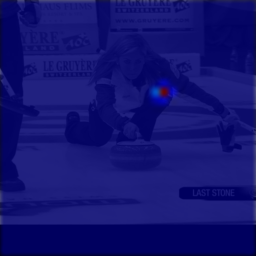}\vspace{0.1mm}\\
    \includegraphics[width=1.96cm]{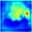}
    \includegraphics[width=1.96cm]{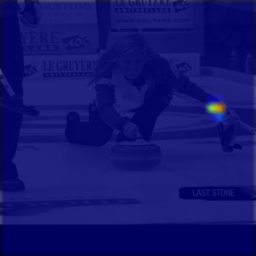}\vspace{0.1mm}\\
    \includegraphics[width=3.99cm]{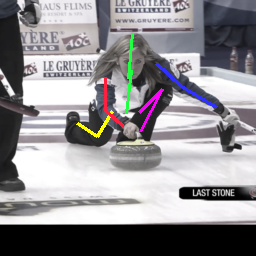}
  \end{subfigure}
  \hspace{0.1mm}
  \caption{
    Indirectly learned part-based heat maps from our method. All the joints
    encoded to RGB are shown in the first image (top-left corner) and the final
    pose is shown in the last image (bottom-right corner). On each column, the
    intermediate images correspond to the predicted heat maps before (left) and
    after (right) the Softmax normalization. The presented heat maps correspond
    to \textit{right ankle},
    \textit{right hip},
    \textit{right wrist},
    \textit{right shoulder},
    \textit{upper neck},
    \textit{head top},
    \textit{left knee}, and
    \textit{left wrist}.
  }
  \label{fig:part-based-maps}
\end{figure}

As suggested in Section~\ref{sec:sam}, the proposed \Softargmax{} function acts
as a constrain on the regression approach, driving the network to learn
part-based detectors indirectly.
This effect provides the flexibility of regression based methods, which can be
easily integrated to provide 2D pose estimation to other applications such as
3D pose estimation or action recognition, while preserving the performance
of detection based methods.
Some examples of part-based maps indirectly learned by our method are show in
Fig.~\ref{fig:part-based-maps}.
As we can see, the responses are very well localized on the true location of
the joints without explicitly requiring so.

\begin{figure*}[hbpt!]
  \centering
  \begin{subfigure}[b]{0.16\textwidth}
    \includegraphics[width=2.9cm]{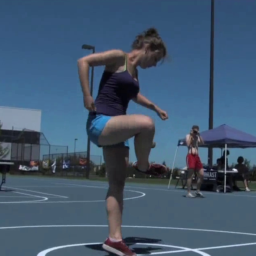}\vspace{0.7mm}\\
    \includegraphics[width=2.9cm]{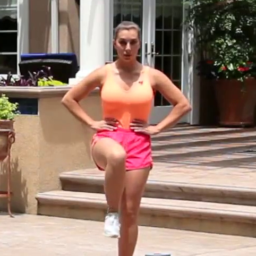}\vspace{0.7mm}\\
    \includegraphics[width=2.9cm]{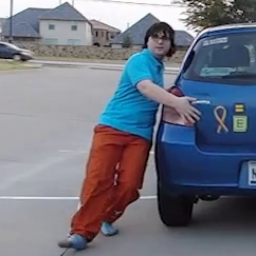}
    \caption{}
  \end{subfigure}
  \begin{subfigure}[b]{0.16\textwidth}
    \includegraphics[width=2.9cm]{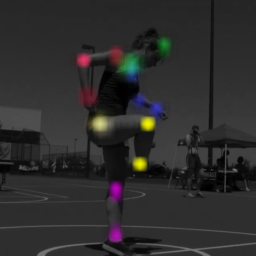}\vspace{0.7mm}\\
    \includegraphics[width=2.9cm]{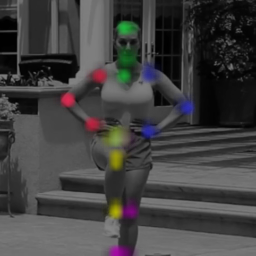}\vspace{0.7mm}\\
    \includegraphics[width=2.9cm]{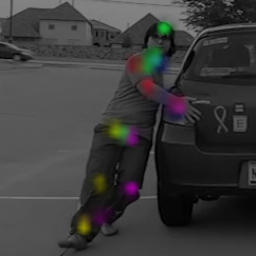}
    \caption{}
  \end{subfigure}
  \begin{subfigure}[b]{0.16\textwidth}
    \includegraphics[width=2.9cm]{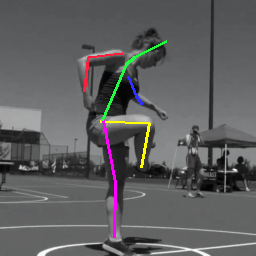}\vspace{0.7mm}\\
    \includegraphics[width=2.9cm]{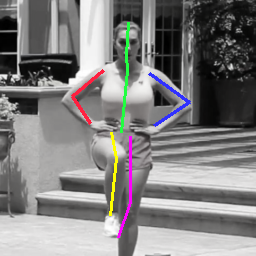}\vspace{0.7mm}\\
    \includegraphics[width=2.9cm]{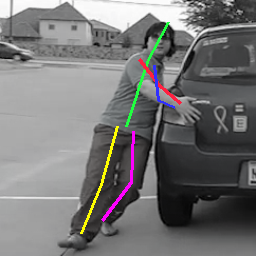}
    \caption{}
  \end{subfigure}
  \begin{subfigure}[b]{0.16\textwidth}
    \includegraphics[width=2.9cm]{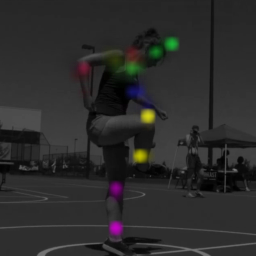}\vspace{0.7mm}\\
    \includegraphics[width=2.9cm]{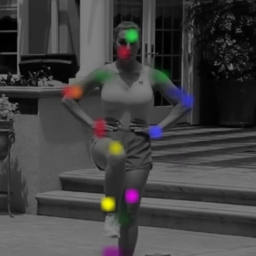}\vspace{0.7mm}\\
    \includegraphics[width=2.9cm]{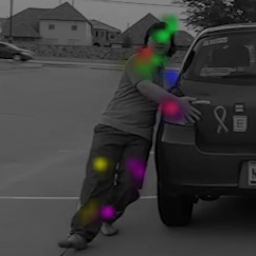}
    \caption{}
  \end{subfigure}
  \begin{subfigure}[b]{0.16\textwidth}
    \includegraphics[width=2.9cm]{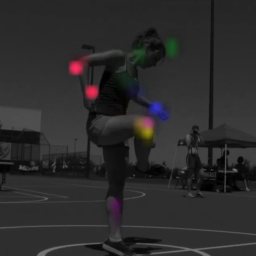}\vspace{0.7mm}\\
    \includegraphics[width=2.9cm]{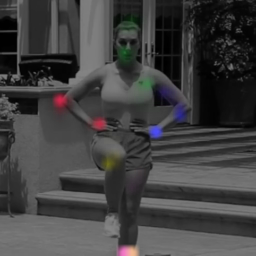}\vspace{0.7mm}\\
    \includegraphics[width=2.9cm]{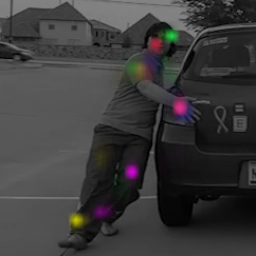}
    \caption{}
  \end{subfigure}
  \begin{subfigure}[b]{0.16\textwidth}
    \includegraphics[width=2.9cm]{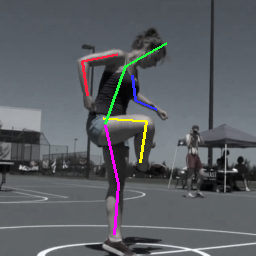}\vspace{0.7mm}\\
    \includegraphics[width=2.9cm]{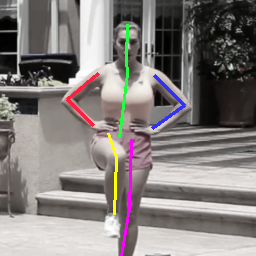}\vspace{0.7mm}\\
    \includegraphics[width=2.9cm]{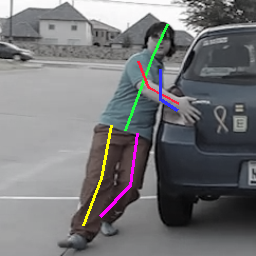}
    \caption{}
  \end{subfigure}
  \caption{
    Samples of context maps aggregated to refine predicted pose. Input image
    (a), part-based detection maps (b), predicted pose without context (c),
    two different context maps (d) and (e), and the final pose with aggregated
    predictions (f).
  }
  \label{fig:context}
\end{figure*}

Additionally to the part-based maps, the contextual maps give extra information
to refine the predicted pose.
In some cases, the contextual maps provide strong responses to regions around
the joint location. In such cases, the aggregation scheme is able to refine the
predicted joint position. On the other hand, if the contextual map response is
weak, the context reflects in very few changes on the pose.
Some examples of predicted poses and visual contributions from contextual
aggregation are shown in Fig.~\ref{fig:context}.
The contextual maps are able to increase the precision of the predictions by
providing complementary information, as we can see for the right elbows of the
poses in Fig.~\ref{fig:context}.

\section{Conclusion}
\label{sec:conclusion}

In this work, we presented a new regression method for human pose
estimation from still images.
The method is based on the \Softargmax{} operation, a differentiable operation
that can be integrated in a deep convolutional network to learn part-based
detection maps indirectly, resulting in a significant improvement over the
state-of-the-art scores from regression methods and very competitive results
compared to detection based approaches.
Additionally, we demonstrate that contextual information can be seamless
integrated into our framework by using additional context maps and joint
probabilities.
As a future work, other methods could be build up to our approach to provide 3D
pose estimation or human action recognition from pose in a fully differentiable
way.

\section*{Acknowledgment}

This work was partially supported by the Brazilian National Council for
Scientific and Technological Development (CNPq) -- Grant 233342/2014-1.

{\small
\bibliographystyle{ieee}
\bibliography{bibliography}

\begin{thebibliography}{10}\itemsep=-1pt

\bibitem{Andriluka_CVPR_2014}
M.~Andriluka, L.~Pishchulin, P.~Gehler, and B.~Schiele.
\newblock {2D Human Pose Estimation: New Benchmark and State of the Art
  Analysis}.
\newblock In {\em {IEEE Conference on Computer Vision and Pattern Recognition
  (CVPR)}}, June 2014.

\bibitem{Andriluka_CVPR_2009}
M.~Andriluka, S.~Roth, and B.~Schiele.
\newblock {Pictorial structures revisited: People detection and articulated
  pose estimation}.
\newblock In {\em Computer Vision and Pattern Recognition (CVPR)}, pages
  1014--1021, June 2009.

\bibitem{Belagiannis_ICCV_2015}
V.~Belagiannis, C.~Rupprecht, G.~Carneiro, and N.~Navab.
\newblock Robust optimization for deep regression.
\newblock In {\em International Conference on Computer Vision (ICCV)}, pages
  2830--2838, Dec 2015.

\bibitem{Belagiannis_2016}
V.~Belagiannis and A.~Zisserman.
\newblock Recurrent human pose estimation.
\newblock {\em CoRR}, abs/1605.02914, 2016.

\bibitem{Bulat_ECCV_2016}
A.~Bulat and G.~Tzimiropoulos.
\newblock {Human pose estimation via Convolutional Part Heatmap Regression}.
\newblock In {\em European Conference on Computer Vision (ECCV)}, pages
  717--732, 2016.

\bibitem{Carreira_CVPR_2016}
J.~Carreira, P.~Agrawal, K.~Fragkiadaki, and J.~Malik.
\newblock Human pose estimation with iterative error feedback.
\newblock In {\em 2016 IEEE Conference on Computer Vision and Pattern
  Recognition (CVPR)}, pages 4733--4742, June 2016.

\bibitem{Chen_NIPS14}
X.~Chen and A.~Yuille.
\newblock Articulated pose estimation by a graphical model with image dependent
  pairwise relations.
\newblock In {\em Advances in Neural Information Processing Systems (NIPS)},
  2014.

\bibitem{ChenSWLY17}
Y.~Chen, C.~Shen, X.~Wei, L.~Liu, and J.~Yang.
\newblock Adversarial posenet: {A} structure-aware convolutional network for
  human pose estimation.
\newblock {\em CoRR}, abs/1705.00389, 2017.

\bibitem{Chollet2016CoRR}
F.~Chollet.
\newblock Xception: Deep learning with depthwise separable convolutions.
\newblock {\em CoRR}, abs/1610.02357, 2016.

\bibitem{chollet2015keras}
F.~Chollet et~al.
\newblock Keras.
\newblock \url{https://github.com/fchollet/keras}, 2015.

\bibitem{ChouCC17}
C.~Chou, J.~Chien, and H.~Chen.
\newblock Self adversarial training for human pose estimation.
\newblock {\em CoRR}, abs/1707.02439, 2017.

\bibitem{chu2016structure}
X.~Chu, W.~Ouyang, H.~Li, and X.~Wang.
\newblock Structured feature learning for pose estimation.
\newblock In {\em Computer Vision and Pattern Recognition (CVPR)}, 2016.

\bibitem{Chu_CVPR_17}
X.~Chu, W.~Yang, W.~Ouyang, C.~Ma, A.~L. Yuille, and X.~Wang.
\newblock Multi-context attention for human pose estimation.
\newblock {\em arXiv preprint arXiv:1702.07432}, 2017.

\bibitem{Dantone_CVPR_2013}
M.~Dantone, J.~Gall, C.~Leistner, and L.~V. Gool.
\newblock {Human Pose Estimation Using Body Parts Dependent Joint Regressors}.
\newblock In {\em {Computer Vision and Pattern Recognition (CVPR)}}, pages
  3041--3048, June 2013.

\bibitem{Fan_2015_CVPR}
X.~Fan, K.~Zheng, Y.~Lin, and S.~Wang.
\newblock Combining local appearance and holistic view: Dual-source deep neural
  networks for human pose estimation.
\newblock In {\em The IEEE Conference on Computer Vision and Pattern
  Recognition (CVPR)}, June 2015.

\bibitem{Felzenszwalb_IJCV_2005}
P.~F. Felzenszwalb and D.~P. Huttenlocher.
\newblock Pictorial structures for object recognition.
\newblock {\em International Journal of Computer Vision}, 61(1):55--79, 2005.

\bibitem{Gkioxari_ECCV_2016}
G.~Gkioxari, A.~Toshev, and N.~Jaitly.
\newblock {Chained Predictions Using Convolutional Neural Networks}.
\newblock {\em European Conference on Computer Vision (ECCV)}, 2016.

\bibitem{He_2016_CVPR}
K.~He, X.~Zhang, S.~Ren, and J.~Sun.
\newblock Deep residual learning for image recognition.
\newblock In {\em The IEEE Conference on Computer Vision and Pattern
  Recognition (CVPR)}, June 2016.

\bibitem{He_CVPR_2016}
K.~He, X.~Zhang, S.~Ren, and J.~Sun.
\newblock {Deep Residual Learning for Image Recognition}.
\newblock In {\em IEEE Conference on Computer Vision and Pattern Recognition
  (CVPR)}, 2016.

\bibitem{Hu_CVPR_2016}
P.~Hu and D.~Ramanan.
\newblock Bottom-up and top-down reasoning with convolutional latent-variable
  models.
\newblock In {\em 2016 IEEE Conference on Computer Vision and Pattern
  Recognition (CVPR)}. 2016.

\bibitem{Insafutdinov_ECCV_2016}
E.~Insafutdinov, L.~Pishchulin, B.~Andres, M.~Andriluka, and B.~Schiele.
\newblock {DeeperCut: A Deeper, Stronger, and Faster Multi-Person Pose
  Estimation Model}.
\newblock In {\em {European Conference on Computer Vision (ECCV)}}, May 2016.

\bibitem{Ionesc_ICCV_2011}
C.~Ionescu, F.~Li, and C.~Sminchisescu.
\newblock {Latent structured models for human pose estimation}.
\newblock In {\em {International Conference on Computer Vision (ICCV)}}, pages
  2220--2227, Nov 2011.

\bibitem{Johnson_BMVC_2010}
S.~Johnson and M.~Everingham.
\newblock Clustered pose and nonlinear appearance models for human pose
  estimation.
\newblock In {\em Proceedings of the British Machine Vision Conference}, 2010.

\bibitem{Kiefel2014}
M.~Kiefel and P.~V. Gehler.
\newblock {\em Human Pose Estimation with Fields of Parts}, pages 331--346.
\newblock Springer International Publishing, Cham, 2014.

\bibitem{Ladicky_CVPR_2013}
L.~Ladicky, P.~H.~S. Torr, and A.~Zisserman.
\newblock Human pose estimation using a joint pixel-wise and part-wise
  formulation.
\newblock In {\em Computer Vision and Pattern Recognition (CVPR)}, 2013.

\bibitem{Lifshitz_ECCV_2016}
I.~Lifshitz, E.~Fetaya, and S.~Ullman.
\newblock {\em Human Pose Estimation Using Deep Consensus Voting}, pages
  246--260.
\newblock Springer International Publishing, Cham, 2016.

\bibitem{Newell_ECCV_2016}
A.~Newell, K.~Yang, and J.~Deng.
\newblock {Stacked Hourglass Networks for Human Pose Estimation}.
\newblock {\em {European Conference on Computer Vision (ECCV)}}, pages
  483--499, 2016.

\bibitem{Ouyang_CVPR_2014}
W.~Ouyang, X.~Chu, and X.~Wang.
\newblock Multi-source deep learning for human pose estimation.
\newblock In {\em Computer Vision and Pattern Recognition (CVPR)}, pages
  2337--2344, June 2014.

\bibitem{pfister2015flowing}
T.~Pfister, J.~Charles, and A.~Zisserman.
\newblock Flowing convnets for human pose estimation in videos.
\newblock In {\em International Conference on Computer Vision (ICCV)}, 2015.

\bibitem{pfister2014deep}
T.~Pfister, K.~Simonyan, J.~Charles, and A.~Zisserman.
\newblock Deep convolutional neural networks for efficient pose estimation in
  gesture videos.
\newblock In {\em Asian Conference on Computer Vision (ACCV)}, 2014.

\bibitem{Pishchulin_CVPR_2013}
L.~Pishchulin, M.~Andriluka, P.~Gehler, and B.~Schiele.
\newblock {Poselet Conditioned Pictorial Structures}.
\newblock In {\em Computer Vision and Pattern Recognition (CVPR)}, pages
  588--595, June 2013.

\bibitem{Pishchulin_ICCV_2013}
L.~Pishchulin, M.~Andriluka, P.~V. Gehler, and B.~Schiele.
\newblock Strong appearance and expressive spatial models for human pose
  estimation.
\newblock In {\em International Conference on Computer Vision (ICCV)}, pages
  3487--3494, 2013.

\bibitem{Pishchulin_CVPR_2016}
L.~Pishchulin, E.~Insafutdinov, S.~Tang, B.~Andres, M.~Andriluka, P.~Gehler,
  and B.~Schiele.
\newblock {DeepCut: Joint Subset Partition and Labeling for Multi Person Pose
  Estimation}.
\newblock In {\em IEEE Conference on Computer Vision and Pattern Recognition
  (CVPR)}, June 2016.

\bibitem{Rafi_BMVC_2016}
U.~Rafi, I.~Kostrikov, J.~Gall, and B.~Leibe.
\newblock An efficient convolutional network for human pose estimation.
\newblock In {\em BMVC}, volume~1, page~2, 2016.

\bibitem{Ramakrishna_ECCV_2014}
V.~Ramakrishna, D.~Munoz, M.~Hebert, A.~J. Bagnell, and Y.~Sheikh.
\newblock {Pose Machines: Articulated Pose Estimation via Inference Machines}.
\newblock In {\em {European Conference on Computer Vision (ECCV)}}, 2014.

\bibitem{Rogez_CVPR_2017}
G.~Rogez, P.~Weinzaepfel, and C.~Schmid.
\newblock {LCR-Net: Localization-Classification-Regression for Human Pose}.
\newblock In {\em {Conference on Computer Vision and Pattern Recognition
  (CVPR)}}, June 2017.

\bibitem{Simonyan15}
K.~Simonyan and A.~Zisserman.
\newblock Very deep convolutional networks for large-scale image recognition.
\newblock In {\em International Conference on Learning Representations}, 2015.

\bibitem{Sun_2017}
X.~Sun, J.~Shang, S.~Liang, and Y.~Wei.
\newblock Compositional human pose regression.
\newblock {\em arXiv preprint arXiv:1702.07432}, 2017.

\bibitem{SzegedyIV16}
C.~Szegedy, S.~Ioffe, and V.~Vanhoucke.
\newblock Inception-v4, inception-resnet and the impact of residual connections
  on learning.
\newblock {\em CoRR}, abs/1602.07261, 2016.

\bibitem{Tompson_CVPR_2015}
J.~Tompson, R.~Goroshin, A.~Jain, Y.~LeCun, and C.~Bregler.
\newblock {Efficient object localization using Convolutional Networks}.
\newblock In {\em {IEEE Conference on Computer Vision and Pattern Recognition
  (CVPR)}}, pages 648--656, June 2015.

\bibitem{Tompson_NIPS_2014}
J.~J. Tompson, A.~Jain, Y.~LeCun, and C.~Bregler.
\newblock Joint training of a convolutional network and a graphical model for
  human pose estimation.
\newblock In Z.~Ghahramani, M.~Welling, C.~Cortes, N.~D. Lawrence, and K.~Q.
  Weinberger, editors, {\em Advances in Neural Information Processing Systems
  27}, pages 1799--1807. Curran Associates, Inc., 2014.

\bibitem{Toshev_CVPR_2014}
A.~Toshev and C.~Szegedy.
\newblock {DeepPose: Human Pose Estimation via Deep Neural Networks}.
\newblock In {\em {Computer Vision and Pattern Recognition (CVPR)}}, pages
  1653--1660, 2014.

\bibitem{Wei_CVPR_2016}
S.-E. Wei, V.~Ramakrishna, T.~Kanade, and Y.~Sheikh.
\newblock {Convolutional pose machines}.
\newblock In {\em {IEEE Conference on Computer Vision and Pattern Recognition
  (CVPR)}}, 2016.

\bibitem{yang2016end}
W.~Yang, W.~Ouyang, H.~Li, and X.~Wang.
\newblock End-to-end learning of deformable mixture of parts and deep
  convolutional neural networks for human pose estimation.
\newblock In {\em Computer Vision and Pattern Recognition (CVPR)}, 2016.

\bibitem{Yang_CVPR_2012}
Y.~Yang, S.~Baker, A.~Kannan, and D.~Ramanan.
\newblock Recognizing proxemics in personal photos.
\newblock In {\em 2012 IEEE Conference on Computer Vision and Pattern
  Recognition}, pages 3522--3529, June 2012.

\bibitem{Yu2016}
X.~Yu, F.~Zhou, and M.~Chandraker.
\newblock {\em Deep Deformation Network for Object Landmark Localization},
  pages 52--70.
\newblock Springer International Publishing, Cham, 2016.

\end{thebibliography}
}

\end{document}